%% file: 00_main.tex
\documentclass[10pt,twocolumn,letterpaper]{article}

\usepackage[pagenumbers]{cvpr} %

\usepackage{graphicx}
\usepackage{amsmath}
\usepackage{amssymb}
\usepackage{booktabs}
\usepackage{xcolor}
\usepackage{tabularx}
\usepackage{wrapfig}
\usepackage{graphicx}
\usepackage{amsmath}
\usepackage{graphicx}
\usepackage{lscape}
\usepackage[title]{appendix}
\usepackage{multicol}

\newcommand{\bR}{\mathbb{R}}

\usepackage{pifont}
\usepackage{amssymb}
\usepackage{xcolor}
\newcommand{\greencheck}{{\color{green}\checkmark}}
\newcommand{\xmark}{\ding{55}}%
\newcommand{\redcross}{{\color{red}\xmark}}
\usepackage[pagebackref,breaklinks,colorlinks]{hyperref}

\usepackage[capitalize]{cleveref}
\crefname{section}{Sec.}{Secs.}
\Crefname{section}{Section}{Sections}
\Crefname{table}{Table}{Tables}
\crefname{table}{Tab.}{Tabs.}

\usepackage{booktabs}%

\def\cvprPaperID{2193} %
\def\confName{CVPR}
\def\confYear{2023}

\begin{document}

\title{A General Purpose Supervisory Signal for Embodied Agents}

\author{Kunal Pratap Singh \and Jordi Salvador \and Luca Weihs \and Aniruddha Kembhavi \\
\vspace{-1.5em}
\and Allen Institute for AI \\
{\tt\small \{kunals, jordis, lucaw, anik\}@allenai.org}
}

\twocolumn[{%
\renewcommand\twocolumn[1][]{#1}%
\maketitle
\begin{center}
    \centering
    \captionsetup{type=figure}
    \vspace{-1em}
    \includegraphics[width=\textwidth]{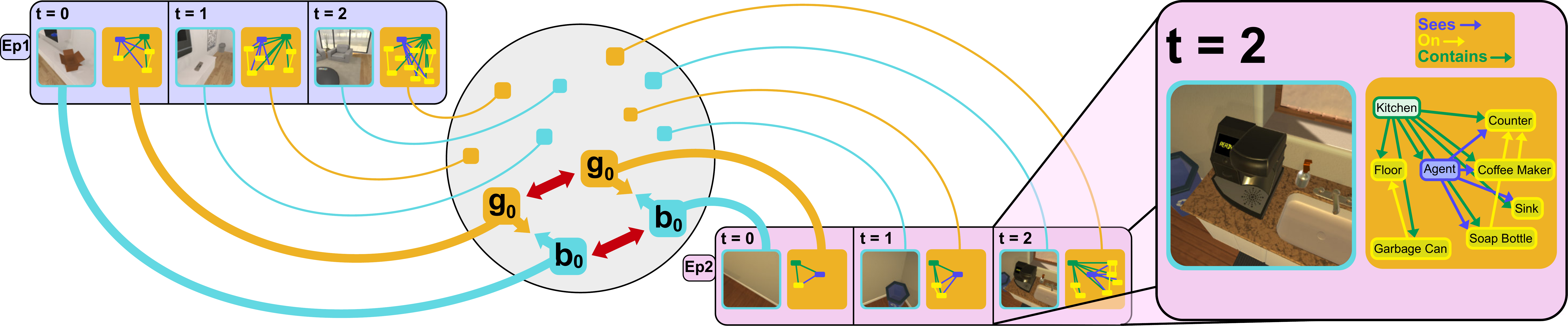}
    \captionof{figure}{\textbf{Scene Graph Contrastive~(SGC) Learning.} We propose to use scene graphs as an auxiliary supervisory signal for embodied agents. We iteratively build a scene graph based on the agent's observations. The agent, the current room, and the objects are represented as nodes in the graph, as shown in the magnified $t=2$ window. Edges of the graph encode various relationships like \emph{Sees}, \emph{On} and \emph{Contains}. 
    We optimize the agent belief to be closer to the graph representation at that time step. Ep1 and Ep2 denote two different episode rollouts. 
    }\label{fig:teaser}
\end{center}%
}]

\begin{abstract}
Training effective embodied AI agents often involves manual reward engineering, expert imitation, specialized components such as maps, or leveraging additional sensors for depth and localization. Another approach is to use neural architectures alongside self-supervised objectives which encourage better representation learning. In practice, there are few guarantees that these self-supervised objectives encode task relevant information. We propose the Scene Graph Contrastive~(SGC) loss, which uses scene graphs as general-purpose, training-only, supervisory signals. The SGC loss does away with explicit graph decoding and instead uses contrastive learning to align an agent's representation with a rich graphical encoding of its environment. The SGC loss is generally applicable, simple to implement, and encourages representations that encode objects' semantics, relationships, and history. By using the SGC loss, we attain large gains on three embodied tasks: Object Navigation, Multi-Object Navigation, and Arm Point Navigation. Finally, we present studies and analyses which demonstrate the ability of our trained representation to encode semantic cues about the environment.

\end{abstract}

\input{01_intro}

\input{02_related}

\input{03_approach}

\input{04_results}
\input{05_conclusion}
{
\small
\bibliographystyle{ieee_fullname}
\bibliography{egbib}
}

\newpage
\appendix
\input{06_appendix}

\end{document}

%% file: 01_intro.tex
\section{Introduction.}

Researchers have pursued designing embodied agents with general neural architectures and training them via end-to-end reinforcement learning (RL) to flexibly complete a range of complex tasks. In practice, however, training agents to perform long horizon tasks using only terminal rewards has been ineffective and inefficient~\cite{Jain2021GridToPix}, particularly in complex visual environments with high-dimensional sensor inputs and large action spaces. This has led to the use of several common ``tricks'' to improve training, \eg manually engineered shaped rewards, use of off-the-shelf vision models to pre-process images, imitation learning with expert trajectories, and the use of special purpose mapping architectures~\cite{chaplot2020object,chaplot2020learning,Weihs2021VisualRR,Jain2021GridToPix,Ehsani2022ObjectDis,Ramrakhya2022HabitatWebLE,Khandelwal2022SimpleBE}.

Looking to reduce the need for such ``tricks'', one promising line of work has looked into training agents with RL and \emph{auxiliary losses} to encourage the production of powerful and useful environment representations. These include self-supervised losses like forward prediction~\cite{Gregor2019ShapingBS}, contrastive predictive coding~\cite{Guo2018NeuralPB}, and inverse dynamics~\cite{Pathak2017CuriosityDrivenEB}, that are task and environment independent. The cost of this generality is that there are no guarantees that the resulting representations will encode task-relevant features like the semantic grounding of objects, which is often key to agent success. Moreover, in practice, these losses tend to work well in video game and gridworld environments but are not hugely effective in more complex visual worlds. On the other hand, supervised auxiliary losses such as disturbance avoidance~\cite{ni2021towards}, depth generation~\cite{Mirowski2017LearningTN} are frequently designed to help with specific tasks, but are not generally useful for new tasks.

In this work we propose the Scene Graph Contrastive (SGC) loss. SGC uses, as its supervisory signal, a non-parametric scene graph that develops and transforms iteratively as the agent interacts with its environment. The agent, objects, and rooms are represented as nodes, agent-object (\eg \emph{Sees} and \emph{Touches}) and object-object (\eg \emph{Contains} and \emph{Above}) relationships are edges and, category and spatial coordinates are represented as node attributes. SGC does not employ any graph decoders which tend to be complex, challenging, and expensive to train. Instead it uses a contrastive learning approach in which the agent must ``pick out'' the graph corresponding to the observations it has seen -- a much simpler learning mechanism which still encourages the agent to develop a graph-aware belief state, see Fig.~\ref{fig:teaser}. Moreover, as we iteratively build the ground truth scene graph in an episode, we naturally generate hard negatives samples as graphs from nearby spatio-temporal states will only be subtly different from one another.

The SGC loss has several desirable characteristics. Firstly, it is \textbf{generally applicable}. The resulting belief representations summarize object semantics, relationships, and history, information that is, intuitively, critical for completing many embodied tasks, and our strong results demonstrate the benefit of these representations. Second, it is \textbf{strongly supervised}. While self-supervised methods for embodied systems are valuable, especially when one wishes to train in the real world, strongly supervised losses are more effective and efficient to optimize. There is now a huge community of researchers invested into building simulated embodied environments~\cite{ai2thor,procthor,Szot2021Habitat2T,Savva2019HabitatAP} and research in simulation-to-real transfer increasingly suggests that training in the real world may not be necessary~\cite{chaplot2020object,robothor,truong2022kin2dyn}. Given this, and as the ground truth data is required \emph{only at training time}, we advocate for using any available supervisory data that may be useful for building powerful representations for the real world. Third, it is \textbf{simple to implement}. SGC does not require designing complex specialized decoders for predicting the scene graph and instead leverages well-studied, graph encoder networks and contrastive losses.

We evaluate the SGC loss by training agents on three complex embodied tasks. These tasks include Object Navigation (ObjectNav)~\cite{robothor} (evaluated across four benchmarks), Multi-Object Navigation (MultiON)~\cite{wani2020multion}, and Arm Point Navigation (ArmPointNav)~\cite{Ehsani2021ManipulaTHORAF}. Across each of these tasks, SGC provides very significant absolute improvements of \textbf{10\%} in ObjectNav, \textbf{9\%} in MultiON and \textbf{3.5\%} in ArmPointNav over models trained with pure RL. We find that these agents learn to represent many semantic cues about their environment, which we show via two studies. First, we demonstrate that SGC trained agents can be quickly fine-tuned to novel goal object categories that were observed previously in their environments but never used as target objects for the task. Second, we present linear probing experiments to study how the SGC loss impacts agents' understanding of free-space and object semantics. We find that the representations learned by the SGC trained agent outperform those of an RL-only trained agent; suggesting that the SGC loss encourages agents to better represent these concepts.

In summary, our contributions include: (1) a proposal to use scene graph as a general purpose supervisory signal for training embodied agents, (2) the formulation of a Scene Graph Contrastive (SGC) loss which avoids the need to use complex graph decoders, and (3) a suite of experimental results which demonstrate that the SGC loss leads to significant performance and sample efficiency gains across multiple embodied tasks.

%% file: 02_related.tex
\section{Related Works.}

\noindent\textbf{Embodied AI in practice.}
The Embodied AI community has been working on several embodied tasks such as navigation~\cite{Batra2020ObjectNavRO,Karkus2021DifferentiableSL,Gan2020LookLA,Maksymets2021THDATH}, instruction following~\cite{Shridhar2020ALFREDAB,Anderson2018VisionandLanguageNI,Krantz2020BeyondTN}, manipulation~\cite{Ehsani2021ManipulaTHORAF,Xiang2020SAPIENAS}, embodied question answering~\cite{Das2018EmbodiedQA,Gordon2018IQAVQ}, and rearrangement~\cite{Weihs2021VisualRR,Batra2020RearrangementAC}. 
Open source simulators~\cite{ai2thor,robothor,Szot2021Habitat2T,li2021igibson20} and benchmarks~\cite{Shridhar2020ALFREDAB,Batra2020RearrangementAC,Weihs2021VisualRR} have enabled tremendous progress on these tasks. Recently, large-scale training~\cite{procthor} and stronger visual backbones~\cite{Khandelwal2022SimpleBE} have shown promising transfer across various environments. However, training performant agents often requires ``tricks'' like manually engineered shaped rewards, imitation learning with expert trajectories~\cite{Jain2021GridToPix,Ramrakhya2022HabitatWebLE}, and use of special purpose mapping modules~\cite{chaplot2020object,chaplot2020learning}, which tend to be task specific. In contrast we propose a general-purpose supervisory signal that encourages agents to learn better representations and show it to be effective across multiple tasks.

\noindent\textbf{Auxiliary Tasks in Reinforcement Learning.}
Auxiliary tasks in tandem with the RL task objective have shown promising results in improving sample efficiency and asymptotic task performance for visual reinforcement learning. These can be supervised tasks that provide external signals like depth maps~\cite{Mirowski2017LearningTN,Gordon2019SplitNetSA,Wijmans2019EmbodiedQA}, game internal states~\cite{Lample2017PlayingFG} and reward prediction~\cite{Jaderberg2017ReinforcementLW}. Various unsupervised/self-supervised auxiliary objectives like auto-encoders~\cite{Lange2010DeepAN,Ha2018RecurrentWM,Yarats2021ImprovingSE}, forward~\cite{Gregor2019ShapingBS} and inverse dynamics~\cite{Pathak2017CuriosityDrivenEB}, spatio-temporal mutual information maximization~\cite{Anand2019UnsupervisedSR,Hjelm2019LearningDR}, contrastive learning~\cite{Guo2018NeuralPB,Guo2020BootstrapLR,Ye2020AuxiliaryTS,Srinivas2020CURLCU}, derive supervision from the agent's own experience. Recently,~\cite{Ye2020AuxiliaryTS,ye2021auxiliary} have shown that self-supervised auxiliary tasks can improve sample efficiency on embodied navigation tasks~\cite{Anderson2018OnEO,Batra2020ObjectNavRO}. 

Unfortunately these approaches may not effectively encode task-relevant features, and therefore fail to provide improvements on complex tasks in photo-realistic environments~\cite{ai2thor,Savva2019HabitatAP}. To alleviate this, we propose to use scene graphs as an auxiliary supervisory signal and show that it leads to more performant agents across different tasks.

\noindent\textbf{Scene Graphs.}
Building rich scene representations has been an active area of research including approaches to build graphs from static images~\cite{dai2017detecting,Li2017SceneGG,lu2016visual,xu2017scenegraph,yang2018graph,zellers2018scenegraphs} and ones that contain temporal information from videos~\cite{cong2021spatial,ji2020action,Liu2020BeyondSS,Ost2021NeuralSG,Shang2017VideoVR,Tsai2019VideoRR}. These methods capture 2D spatial relationships between objects. There has also been work that aims to encode 3D relationships~\cite{Armeni20193DSG,Chen2019HolisticSU,Fisher2011CharacterizingSR,Wald2020Learning3S,Zhou2019SceneGraphNetNM}.

Scene graphs have been used in embodied settings for learning physics engines~\cite{Battaglia2016InteractionNF}, visual navigation~\cite{Du2020LearningOR,Yang2019VisualSN,Chaplot2020NeuralTS,Chen2019ABA,Savinov2018SemiparametricTM,Wu2019BayesianRM}, manipulation~\cite{Zhu2021HierarchicalPF} and building actionable representations~\cite{Li2021EmbodiedSS,Rosinol20203DDS}. 
There has also been work~\cite{gadre2022csr} to encode scene graph relationships for use in downstream tasks like object tracking and room rearrangement. 
However, contrary to this stream of work, we do not attempt to generate scene graphs or use it as an input to the agent. Instead, we use it as an auxiliary training signal in a contrastive learning setup. This avoids using complex graph decoding while enabling graph-aware belief representations.

%% file: 03_approach.tex
\section{Approach.}
\label{sec:approach}
In this section we present the Scene Graph Contrastive~(SGC) loss for aiding embodied agent training. We begin by discussing the approach of iteratively building a scene graph from the agent's observations. Subsequently, we discuss how we use scene graphs as a training signal in our contrastive learning framework. Lastly, we describe our model architecture and how we train embodied agents for various tasks with our proposed loss.
\begin{figure}[t!]
    \includegraphics[width=0.5\textwidth]{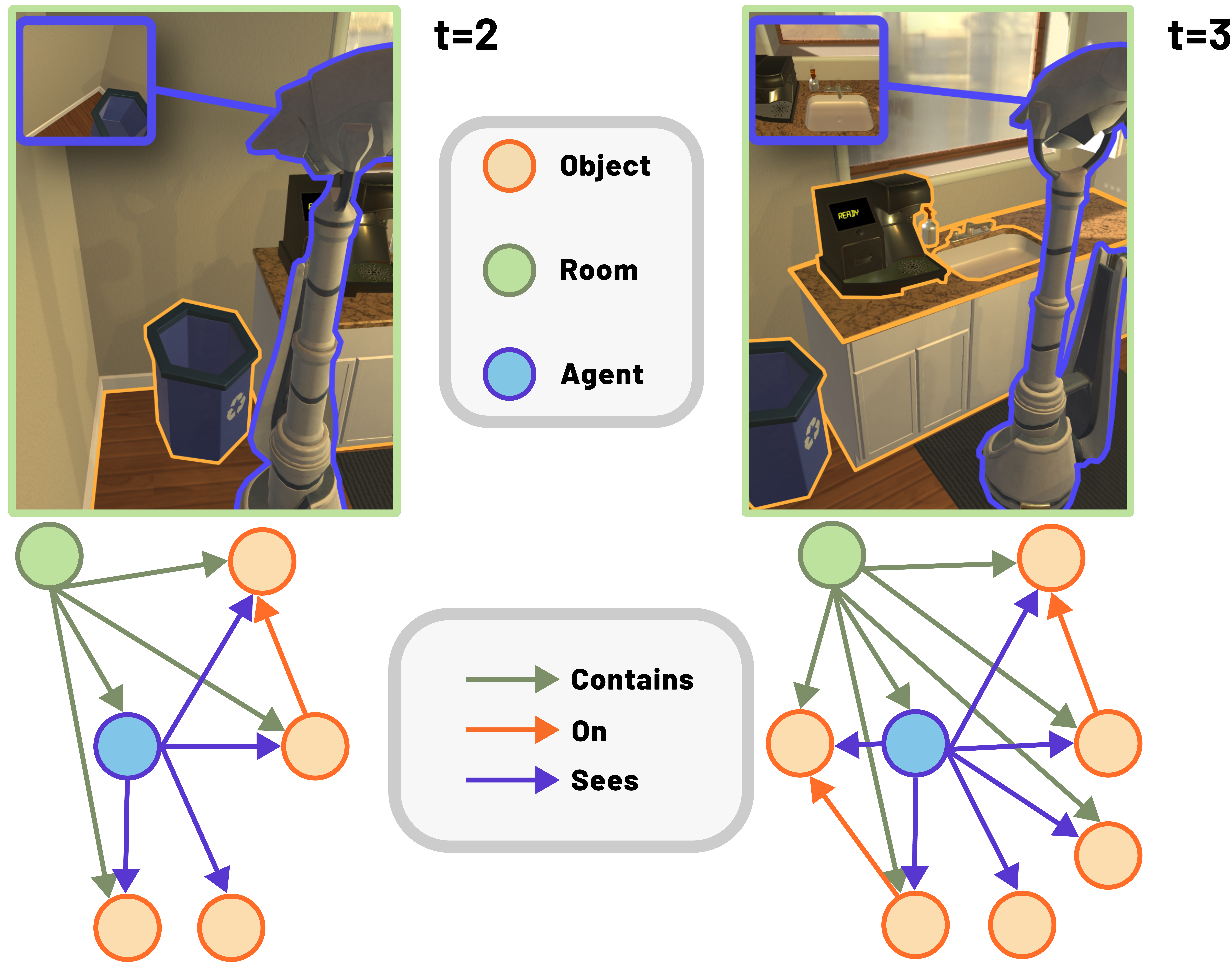}
    \caption{\textbf{Iterative Graph Building.} Illustrative example of how a scene graph is built. The agent, the room, and objects are all added as nodes to the graph. As the agent moves in the environment, we add the objects that it sees, for instance at t=3, the coffee machine, sink, and counter top. We also add edges that signify various relationships like \emph{Contains}, \emph{On}, and \emph{Sees}.}
    \vspace{-2em}
    \label{fig:graph_build}
\end{figure}

\begin{figure*}[ht!]
    \includegraphics[width=\textwidth]{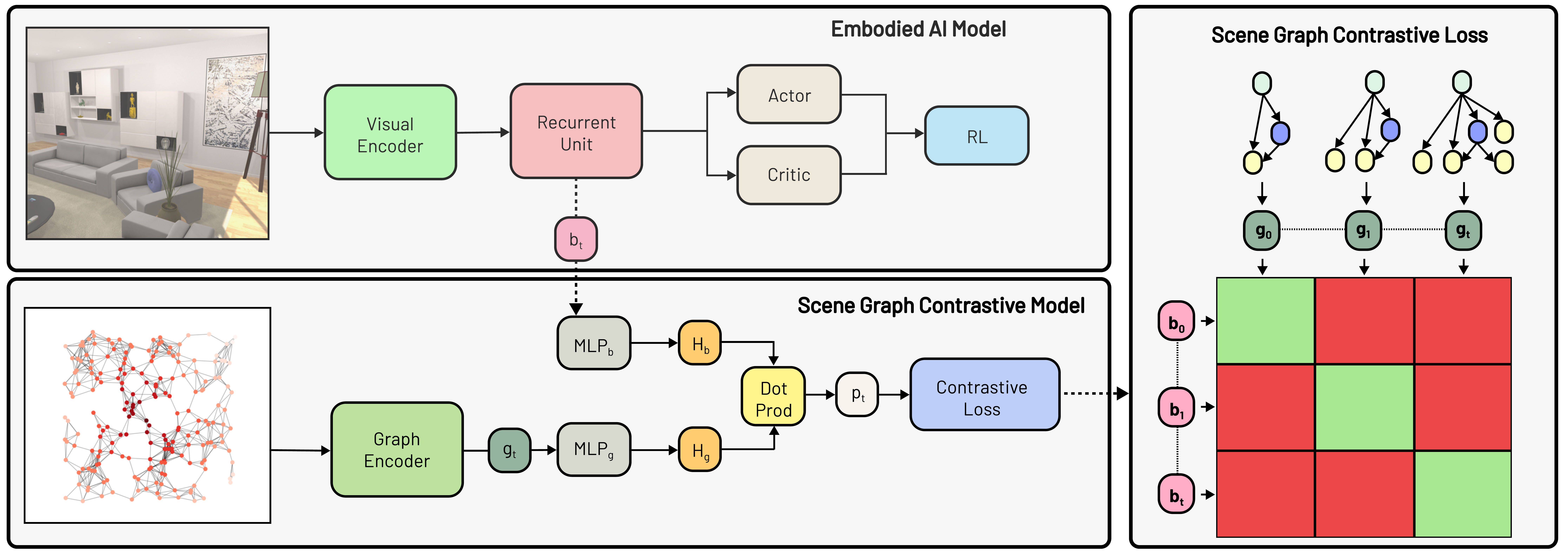}
    \caption{\textbf{Scene Graph Contrastive~(SGC) framework.} 
    We show a standard Embodied-AI model which takes, as input, an egocentric RGB observation and outputs a policy and scalar value for RL training.
    We supplement this with our Scene Graph Contrastive Model. It takes as input, the belief, $b_t$, and graph representation $g_t$, at each time step. The prediction of this model, $p_t$ is passed into the Scene Graph Contrastive Loss. The SGC Loss optimizes the model to predict which graph embedding belongs to a particular time step, $t$. 
    }
    \label{fig:model}
\vspace{-1.5em}    
\end{figure*}

\subsection{Iterative Scene Graph.}
\label{sec:scene_graph_build}
Embodied AI simulators provide a rich trove of information like scene semantics, object positions, geometry, and spatial relationships. We attempt to distill this information by building a scene graph consistent with the agent's explored environment and use it to construct a supervisory signal. 
We define this scene graph as a non-parametric, object-centric, directed, graph representation. 
Figure~\ref{fig:graph_build} shows an example of how we iteratively build the scene based on the agent's exploration in the environment.

\noindent\textbf{Node Features.}
We build a scene graph that iteratively updates based on the agent's path through the environment. We begin with the agent as the first node in the graph. Then, we add all the objects as nodes that are visible and within a threshold distance of the agent. Every instance of a particular object type is treated as a separate node. Additionally, if we're operating in house-sized environment, we add the room in which the agent is currently present as a node in the graph. 
Note that, once an object node is added in the graph, it continues to exist on the subsequent time step graphs as well, even if the object goes out of view. 
This allows the scene graph to retain the history of an episode, which can be a useful attribute for long-horizon tasks.

Additionally, we also assign node-specific features to each node. These comprise of a concatenation of (1) an embedding of the object's type and, (2) the $(x, y, z)$ 3-D coordinates of the object, \ie its position. Each object position is defined relative to the agent, to encode spatial awareness about the environment with respect to the agent's current state. Note that these node features are updated after every agent step as they would otherwise quickly become invalid.

\noindent\textbf{Edge Features and Relationships.}
We use the edges of the scene graphs to encode various relationships between the nodes.
These relationships can be Agent-Object like \emph{Sees}, \emph{Holds} or \emph{Touches}. Other agent-agnostic relationships include Object-Object relationships like \emph{On} or \emph{Near}. We also have Agent conditioned Object-Object relationships like \emph{Right}, \emph{Left}, \emph{Above} which depend on the object positions relative to the agent perspective.
We provide a list of all relationships and how they are estimated in the supplementary materials. We also have a relationship \emph{Contains}, between the rooms of a house and other nodes, that determine whether the agent or a particular object is present in that room.
At each time step, we compute these relationships between all the nodes based on their positions and geometry. The information needed to compute these relationships, i.e.\ object and agent poses, is readily available in most open source simulators~\cite{ai2thor,robothor,procthor,Szot2021Habitat2T,Shen2021iGibson1A,Xiang2020SAPIENAS,Gan2021ThreeDWorld}, making it straightforward to construct this scene graph.

If a relationship is true, for instance, \emph{Sees}(Agent, Apple), we add a directed edge between the agent node and apple node with the attribute \emph{Sees} set as true. 
Relationships are updated at each time step. This means that the set of edges between nodes is not static and may change between agent steps, \eg the agent might lift an apple off of a table resulting in the apple no longer being \emph{On} the table.

\subsection{Scene Graph Contrastive Learning.}
\label{sec:loss_motive}
The scene graph described in the Section~\ref{sec:scene_graph_build} is a rich source of information about the environment and can allow the agent to perform a range of tasks with ease. However, the privileged metadata information required for the scene graph's construction is often unavailable during inference time when deploying to unseen environments or real-world settings. Therefore we propose to use it as an auxiliary \emph{training-only} signal. This alleviates its need during inference, allowing us to deploy our trained policies even in the absence of privileged metadata information.

One common approach for leveraging supervisory information for representation learning is simply to build a decoder module that directly attempts to predict that supervisory information. Building such a decoder in our setting is computationally expensive and cumbersome: attempting to directly predict a scene graph with an unknown number of nodes would, similarly as for language prediction, require an iterative decoding mechanism which would, one by one, add nodes and edges to a graph until some stopping criterion was reached. While directly predicting such a graph would certainly have advantages, its computational cost makes it unappealing in a RL setting. We consider another approach.

Intuitively, we would like an agent's representation of their environment, commonly called the agent's belief, to be similar
to a scene graph representation of the agent's environment. We achieve this by training a contrastive loss that effectively asks the agent to \emph{pick out} the scene graph corresponding to its observations at particular time step from among other distractor graphs. Unlike decoding, encoding a scene graph is a significantly easier task, see Sec.~\ref{sec:model_and_loss}.

We collect scene graphs from parallel agent rollouts, which implies that we have some scene graphs in the batch that are from the same episode, and some from entirely different episodes.
As discussed in Section~\ref{sec:scene_graph_build}, we iteratively build the scene graphs as the agent observes new objects in an episode. This enables us to automatically generate hard negative samples as graphs from nearby time steps of the same episode differ very slightly from one another. 
\subsection{Training Embodied Agents with SGC.}
\label{sec:model_and_loss}
We propose to train Embodied-AI models using scene graphs as auxiliary supervision as described in Section~\ref{sec:loss_motive}. As shown in Figure~\ref{fig:model}, we consider a typical Embodied-AI model. It consists of a visual encoder to process the observations from the environment. Following~\cite{Khandelwal2022SimpleBE} we use a frozen CLIP-ResNet50 encoder for encoding our visual observations. We also have a recurrent unit, specifically a GRU for keeping a memory of these visual features, followed by a linear actor-critic layer for reinforcement learning. We refer to the GRU output as beliefs, denoted by $b_t$, where $t$ is the time step.

To enable auxiliary learning we propose a \textbf{Scene Graph Contrastive Model} as shown in Figure~\ref{fig:model}. It consists of a graph encoder, comprised of three Graph Attention Layers~\cite{Velickovic2018GraphAN}, followed by a global max-pool across node features. The graph encoder produces a representation, $g_t$, for the current time step's scene graph.
Following~\cite{Chen2020ASF}, we use two multi-layer perceptrons, denoted by $\textsc{mlp}_b$ and $\textsc{mlp}_g$ in Figure~\ref{fig:model}, to encode the beliefs, namely we let:
\begin{align*}
H_{b_t} &= \textsc{mlp}_b(b_t) \in\bR^D, &H_b = [H_{b_1}\ \ldots\ H_{b_T}] 
\in\bR^{D\times H} \;, \nonumber\\
H_{g_t} &= \textsc{mlp}_g(g_t) \in\bR^D, &H_g = [H_{g_1}\ \ldots\ H_{g_H}]\in\bR^{D\times H} \;.
\end{align*}
Without loss of generality, above we consider the case where our loss is computed using $H$ sequential agent steps; in practice, our loss will be computed on a batch of such trajectories.
As shown in Figure~\ref{fig:model}, we take a dot product of embeddings $H_b$ and $H_g$, to generate a prediction matrix $P$:
\begin{align*}
\begin{split}
p_{t,s} = H_{b_t} \cdot H_{g_s},\quad P = H_b^T H_g = [p_{t,s}] \in \bR^{H\times H}.
\end{split}
\end{align*}

We take a softmax across the columns of P, and pass each row to a cross entropy loss. The ground truth for this loss is a diagonal matrix as shown in the Scene Graph Contrastive~(SGC) Loss panel in Figure~\ref{fig:model}, \ie the entries of $P$ where $t=s$. This classification objective attempts to predict which graph embedding belongs to a particular time step. Note that the SGC loss, is optimized as an auxiliary objective alongside usual reinforcement learning losses; in our experiments we use the DD-PPO RL loss~\cite{Wijmans2020DDPPO}.

It is worth noting that, constructing a scene graph at every step can be computationally expensive. To avoid this overhead, we randomly sample time steps at which we generate the graph, and only compute the loss at those steps. We show the effectiveness of SGC through various experiments and analysis in Section~\ref{sec:results}.

%% file: 04_results.tex
\section{Results.}
\label{sec:results}

\begin{table*}[h!]
\centering

\includegraphics[width=0.8\textwidth]{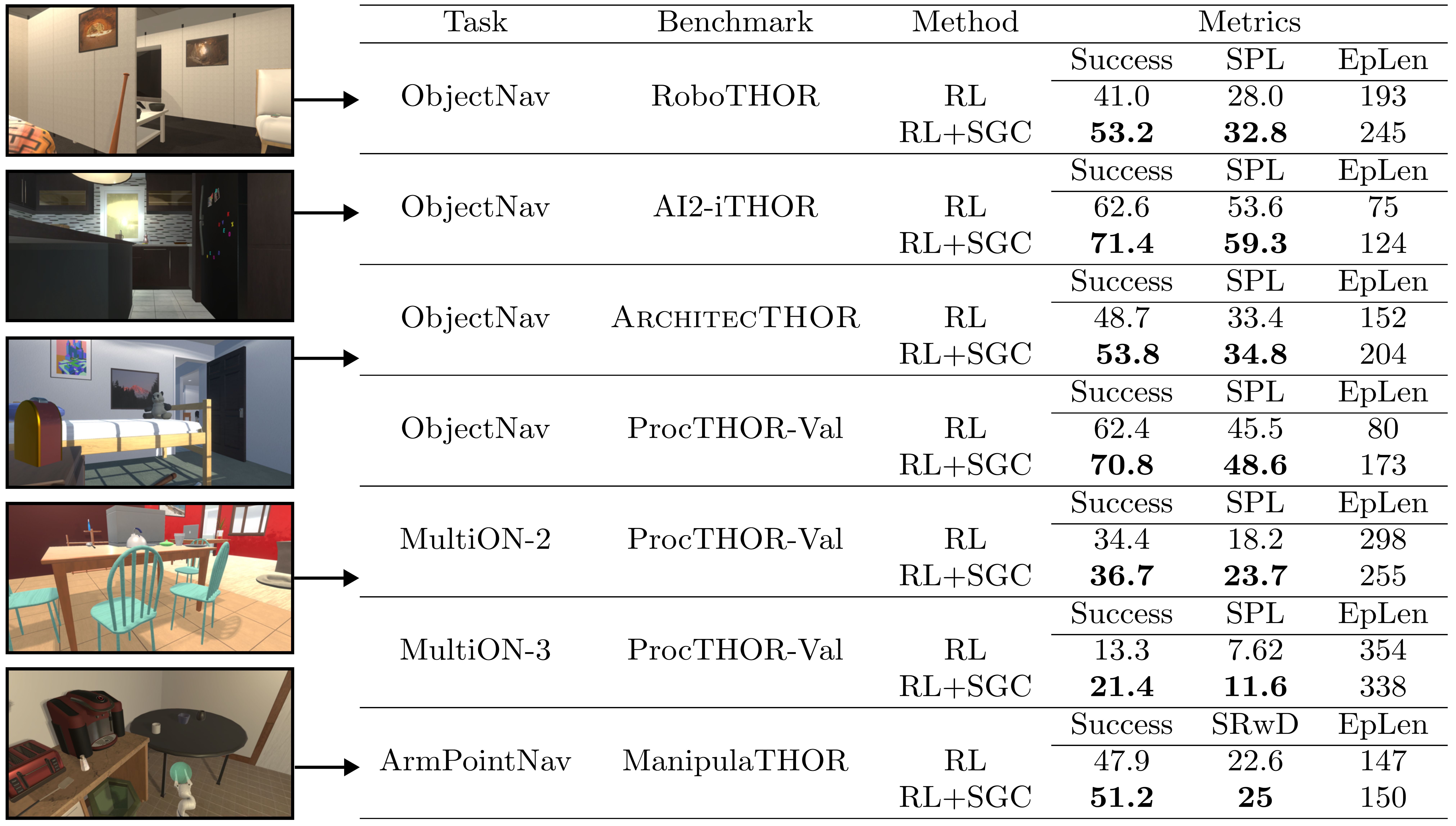}
\caption{\textbf{Results on various Embodied-AI benchmarks.} EpLen represents the average episode length of various models. We train all our models on \emph{ProcTHOR-Train} houses.~\emph{ProcTHOR-Val} represents ProcTHOR validation houses, that the agent never sees during training. Note that our trained agents are evaluated \textbf{zero-shot} on all tasks in the RoboTHOR, AI2-iTHOR, \textsc{ArchitecTHOR}, and ManipulaTHOR benchmarks.
To give an overview of the visual diversity of various benchmarks, we present an egocentric view from each. 
}
\vspace{-1.5em}
\label{tab:results}
\end{table*}

\subsection{Experiment Setup.}
\noindent\textbf{Dataset.} We use the ProcTHOR~\cite{procthor} framework to train agents for various embodied tasks. ProcTHOR provides 10,000 training houses, which we refer as \emph{ProcTHOR-Train}. We train all our agents on these environments. 
ProcTHOR also provides \emph{ProcTHOR-Val}, 1000 validation houses that the agent does not see during. We use the AllenAct~\cite{AllenAct} framework to train our models.

\noindent\textbf{Model variants.} For each task, we train two agents: 

\noindent $\bullet$ \emph{RL}:  This agent is trained with pure reinforcement learning~(RL), specifically DD-PPO~\cite{Wijmans2020DDPPO}.

\noindent $\bullet$ \emph{RL${+}$SGC}: To demonstrate the efficacy of our Scene Graph Contrastive~(SGC) loss, we train this agent with SGC as an auxiliary objective to RL as described in Section~\ref{sec:model_and_loss}.

Note that both agents are trained with the \emph{same} hyperparameter setups. We provide the hyperparameter and training details in the appendix. 
We train these agents on three embodied tasks, namely Object Navigation, Multi-Object Navigation, and Arm Point Navigation. 
We use a frozen CLIP-ResNet50 as our visual encoder, and a GRU as the recurrent unit for all these tasks.
We train our Object Navigation, Multi-Object Navigation, and ArmPointNav models for 350M, 180M, and 90M steps, respectively. 

Additionally, we remark that due to recent updates in AI2-THOR~\cite{ai2thor}, the simulated LoCoBot agent is now allowed to look down by up to 60$^\circ$ (in previous versions this was set to 30$^\circ$). The models presented in~\cite{procthor} were trained before this update, therefore we use the authors' code to retrain their agents and present updated numbers. Note that, for a given task, we use the same hyperparameters for all the models trained on it for fair comparison.
 
\subsection{Object Navigation.}
\noindent\textbf{Task.}
Object Navigation~(ObjectNav) requires an agent to locate a specified object category. The agent begins the episode at a random location and is given a target object category, for instance, apple. The action space consists of \textsc{MoveAhead}, \textsc{RotateLeft}, \textsc{RotateRight}, \textsc{LookUp}, \textsc{LookDown}, and \textsc{End}.
All our ObjectNav agents are trained with a simulated LoCoBot~(Low Cost Robot)~\cite{locobot}, and use egocentric RGB images as input.
The rotation degree for \textsc{Rotate} and \textsc{Look} actions is 30$^\circ$.

\noindent\textbf{Metrics.} 
An episode is considered successful if the agent takes an \textsc{End} action and the target object category is visible and within 1m of the agent.
We report the success rate~(SR) and Success weighted by path length~(SPL)~\cite{Anderson2018OnEO} for this task.

\noindent\textbf{Results.} In Table~\ref{tab:results}, we evaluate our trained models across 4 ObjectNav datasets. To reiterate, we train our models on \emph{ProcTHOR-Train}. First, to show in-domain generalization, we evaluate both agents on \emph{ProcTHOR-Val} and achieve an improvement of \textbf{12\%} in SR and \textbf{4.8\%} in SPL.

Moreover, following~\cite{procthor}, to investigate the cross-domain generalization of our approach, we perform \emph{zero-shot} evaluations on the RoboTHOR, AI2-iTHOR, and \textsc{ArchitecTHOR} ObjectNav datasets. \emph{Zero-shot} here implies, that neither of the models have been trained on scenes from these datasets.
As shown in Table~\ref{tab:results}, we observe that using SGC provides a clear gain across all domains, thereby indicating its effectiveness in producing generally performant ObjectNav models. We see a substantial improvement of \textbf{12\%} on RoboTHOR, \textbf{9\%} on AI2-iTHOR, and \textbf{5\%} on \textsc{ArchitecTHOR} in the SR metric.

Another interesting insight from our experiments is that ObjectNav models trained with SGC consistently traverse much longer trajectories, as reflected by the EpLen metric in Table~\ref{tab:results}. On further investigation, we found that this can be attributed to our RL${+}$SGC agent producing fewer false positives by preferring to wait for episodes to timeout instead of taking the \textsc{End} action when the target object is not visible.
This implies that an agent trained with SGC keeps exploring its environment unless its very certain it has found the target or it exhausts the maximum number of steps allowed.
For instance, in RoboTHOR, we find that agents trained with just RL execute the \textsc{End} action incorrectly in \textbf{45\%} episodes. On the other hand, RL${+}$SGC does so in only \textbf{14.9\%} of the episodes. Our conjecture is that the SGC loss enriches the agent's understanding of the environment and prevents it from misrecognizing objects and, thereby, pre-maturely ending episodes. We observe this trend across all ObjectNav datasets.

\subsection{Multi-Object Navigation.}
\noindent\textbf{Task.} We implement the Multi-Object Navigation~(MultiON) task originally proposed in~\cite{wani2020multion} in ProcTHOR~\cite{procthor} environments. We create two variants, MultiON-2 and MultiON-3, which require the agent to navigate to 2, and 3, objects in an episode respectively. The action space contains \textsc{MoveAhead}, \textsc{RotateLeft}, \textsc{RotateRight}, \textsc{LookUp}, \textsc{LookDown} and \textsc{Found} actions. 
We provide the first goal object category to the agent at the beginning of the episode. Once an agent successfully finds the first target, by calling the \textsc{Found} action with the object visible and nearby, we provide the next target object. 
Similar to ObjectNav, we train our agents with a simulated LoCoBot~\cite{locobot}, and use egocentric RGB images as input. In alignment with a real LoCoBot, the rotation degree for \textsc{Rotate} and \text{Look} actions is 30 degrees. 

\noindent\textbf{Metrics.}
The target object must be visible and within 1m of the agent for the \textsc{Found} action to be successful.
An episode is successful if the agent can finds all the target objects. It is considered a failure otherwise.
Following~\cite{wani2020multion}, we report the Success Rate and SPL metrics for this task. 

\noindent\textbf{Results.}
We collect a validation dataset in the \emph{ProcTHOR-Val} environments. We use 200 houses that the agent has \emph{never} seen during training. As mentioned before, we present results on two MultiON variants, MultiON-2 and MultiON-3, where the agent needs to navigate to 2 and 3 target objects respectively. 
As shown in Table~\ref{tab:results}, for MultiON-2, we observe an improvement of \textbf{2.3\%} in Success Rate, and \textbf{5.5\%} in SPL.
We replicate a similar setup for MultiON-3, and see a large \textbf{8\%} improvement in Success Rate, and \textbf{4\%} in SPL.

\subsection{Arm Point Navigation.}
\noindent\textbf{Task.} To evaluate our approach on a manipulation based task, we train models to complete ArmPointNav, a visual mobile manipulation task proposed in~\cite{manipulathor}. This task requires an arm-equipped agent to move a target object from a its starting location to a given goal location. These locations are given to the agent in the agent's relative coordinate frame. Note that we do not use any other visual inputs besides egocentric RGB images.
The action space for the arm agent includes three navigation actions  (\textsc{MoveAhead}, \textsc{RotateLeft}, and \textsc{RotateRight}), with rotations of 45 degrees. It also has 8 arm-based actions, namely \textsc{Move-arm-\{x,y,z\}-\{p,m\}}, which allow the wrist to move in a plus~(\textsc{p}) or minus~(\textsc{m}) direction relative to the agent along the~\textsc{(x,y,z)} axis, and \textsc{Move-Arm-Height-\{p,m\}}, which modifies the arm height.

\noindent\textbf{Metrics.}
An episode is considered successful if the target object reaches the goal location.
We report two metrics, Success Rate~(SR) and Success Rate without Disturbance~(SRwD). SRwD indicates how often the agent can complete the task without colliding with non-target objects.

\noindent\textbf{Results.} 
We evaluate on the AI2-iTHOR test tasks from~\cite{manipulathor}, and report performance for both our models. We see a gain of \textbf{3.3\%} on SR and \textbf{2.4\%} on SRwD. 

\subsection{Ablation and Analysis.}
\subsubsection{SGC \emph{v.s{.}} Other Auxiliary Objectives}
As shown in Table~\ref{tab:results}, training with the SGC loss as an auxiliary objective with RL improves performance across various embodied tasks. However, to investigate how it compares to other auxiliary objectives, we present a comparison with two additional baselines:

\noindent $\bullet$ \emph{RL${+}$CPCA-16}~\cite{Guo2018NeuralPB}: A self-supervised objective that has shown sample efficiency improvements in PointNav~\cite{Ye2020AuxiliaryTS}.

\noindent $\bullet$ \emph{RL${+}$Visibility}: We implement an auxiliary loss in which the agent must predict whether a set of objects are visible or not at a every time step. This supervisory loss can be considered task-specific to ObjectNav as it directly informs the agent whether its seeing a certain object or not, intuitively a strong signal for this task. 

We train these baselines with the same hyperparameter setup as the models presented in Table~\ref{tab:results}. 
Table~\ref{tab:aux_loss_ablation} presents evaluation results on the \emph{ProcTHOR-Val} ObjectNav benchmark. We find that RL${+}$Visibility actually performs worse than RL, meaning that adding an auxiliary loss does not necessarily lead to performance benefits. We suspect that, as objects of most categories will not be visible to the agent at a given time step, the visibility loss is overwhelmed by negative examples and thus fails to provide a strong supervisory signal. This emphasizes the challenge of designing good auxiliary losses: intuition often fails.

RL${+}$CPCA-16 outperforms RL and RL${+}$Visibility, but still lags RL${+}$SGC by 4\% in success rate. Our SGC loss is computed at only 20\% of the timesteps for the purpose of computational efficiency. On the other hand, we computed the RL${+}$CPCA-16 loss at every time step without subsampling; despite this, SGC outperforms.

\begin{table}[t!]
\centering
\begin{tabular}{lcc}
\hline
Model                & SR   & SPL  \\ \hline
RL ${+}$ SGC (ours)        & 70.8 & 48.6 \\
RL ${+}$ CPCA-16    & 66.2     & 45.9    \\ 
RL ${+}$ Visibility &  54.8    & 40.3     \\ 
RL              & 62.4   & 45.5   \\ \hline
\end{tabular}
\caption{\textbf{Comparing SGC with other auxiliary losses.} SR and SPL indicate the success rate and success weighted by path length on the \emph{ProcTHOR-Val} Object Navigation benchmark.}
\vspace{-1.5em}
\label{tab:aux_loss_ablation}
\end{table}
\subsubsection{Adapting to Novel Object categories.}
Today's ObjectNav agents, including the models presented in this work, are trained to find a fixed set of object categories. However, in practice, we may wish to adapt our agents so as to enable them to navigate to novel object types outside their existing vocabulary. One, brute force, solution is to simply retrain from scratch every time we're presented with a new set of object categories. This would require an vast amount of compute and time, making it unfeasible.

We present an alternative solution, we take an ObjectNav model, trained on a set of object categories, and attempt to quickly finetune it on a set of new object categories. We achieve this by freezing the recurrent unit, usually a GRU, thereby preserving the belief representation. After freezing, the only parameters being optimized are the Actor-Critic head, and target object type encoder. This method builds upon the intuition that, after training the belief representations once, the GRU should learn to summarize information about the environment into actionable representations. Therefore, the success of this approach is dependent on, and hence indicative of, the quality of the belief representations. 

To analyze the quality of the learned belief representations of our agents, we finetune them to navigate to novel object categories that they were not previously used as goal objects. Note that these new categories were, however, present in the training environments. We follow the methodology described above and keep the belief representation frozen. 
We sample 5 object types randomly that were not in the initial training object categories, and then fine-tune both models, RL and RL ${+}$ SGC, for 2 million steps in \emph{ProcTHOR-train} environments. We fine-tune both models, with just reinforcement learning~(DD-PPO~\cite{Wijmans2020DDPPO}). To evaluate these fine-tuned models, we collect a validation dataset in the \emph{ProcTHOR-Val} environments with the newly sampled target object categories.
\begin{figure}[t]
\centering
\includegraphics[width=0.5\textwidth]{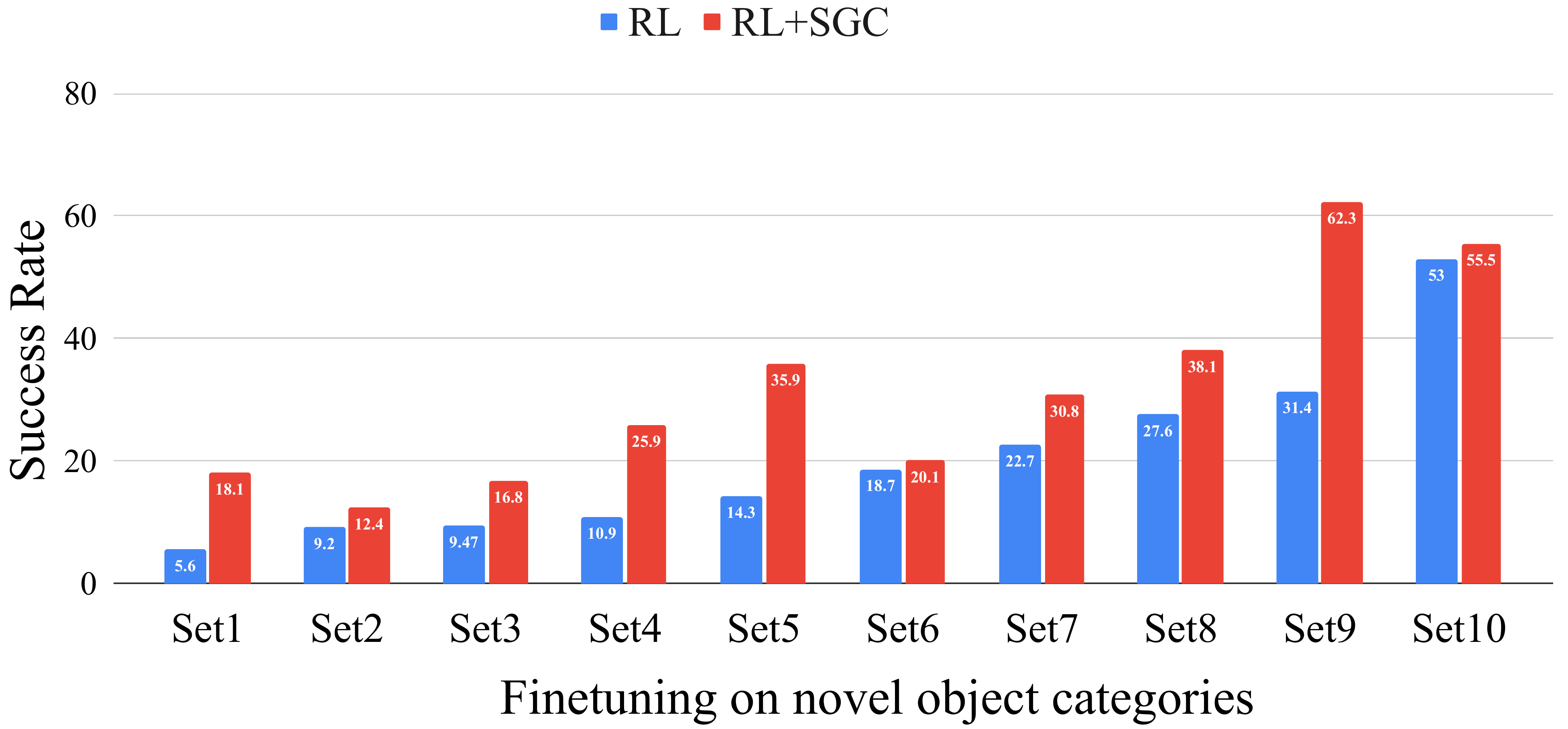}
\caption{\textbf{Adapting to Novel Object types.} We sample 10 sets of 5 novel object types, and finetune RL ${+}$ SGC and RL initialized models with DD-PPO~\cite{Wijmans2020DDPPO} for each set for 2 million steps. We observe consistent improvement on the success rate across all sets. For visualization purpose, we arrange the sets in the ascending order of success for the RL initialized model.}
\vspace{-1.5em}
\label{fig:new_obj_category}
\end{figure}

Figure~\ref{fig:new_obj_category} displays the validation-set success rates of fine-tuned models corresponding to 10 randomly sampled sets of 5 object categories.
We also conduct a paired t-test to ensure that the large observed difference between RL${+}$SGC and RL is statistically significant. We indeed find that the gains in SR and SPL are significant at 0.01 and 0.05 levels, respectively.
Note that RL${+}$SGC initialized models generalize better to objects that are both easy~(Set1), and hard~(Set10), to navigate to. 
The results indicate that the belief representation trained with SGC is able to encode general semantic information about the environment, allowing the SGC model to generalize to novel object categories much faster than the RL model. 
We provide the 10 sets of objects, and the pool that they were sampled from, in the supplementary materials.

\definecolor{thegreen}{RGB}{44,160,40}
\definecolor{theorange}{RGB}{255,127,14}
\definecolor{theblue}{RGB}{31,119,180}
\begin{figure}[t!]
\includegraphics[width=0.5\textwidth]{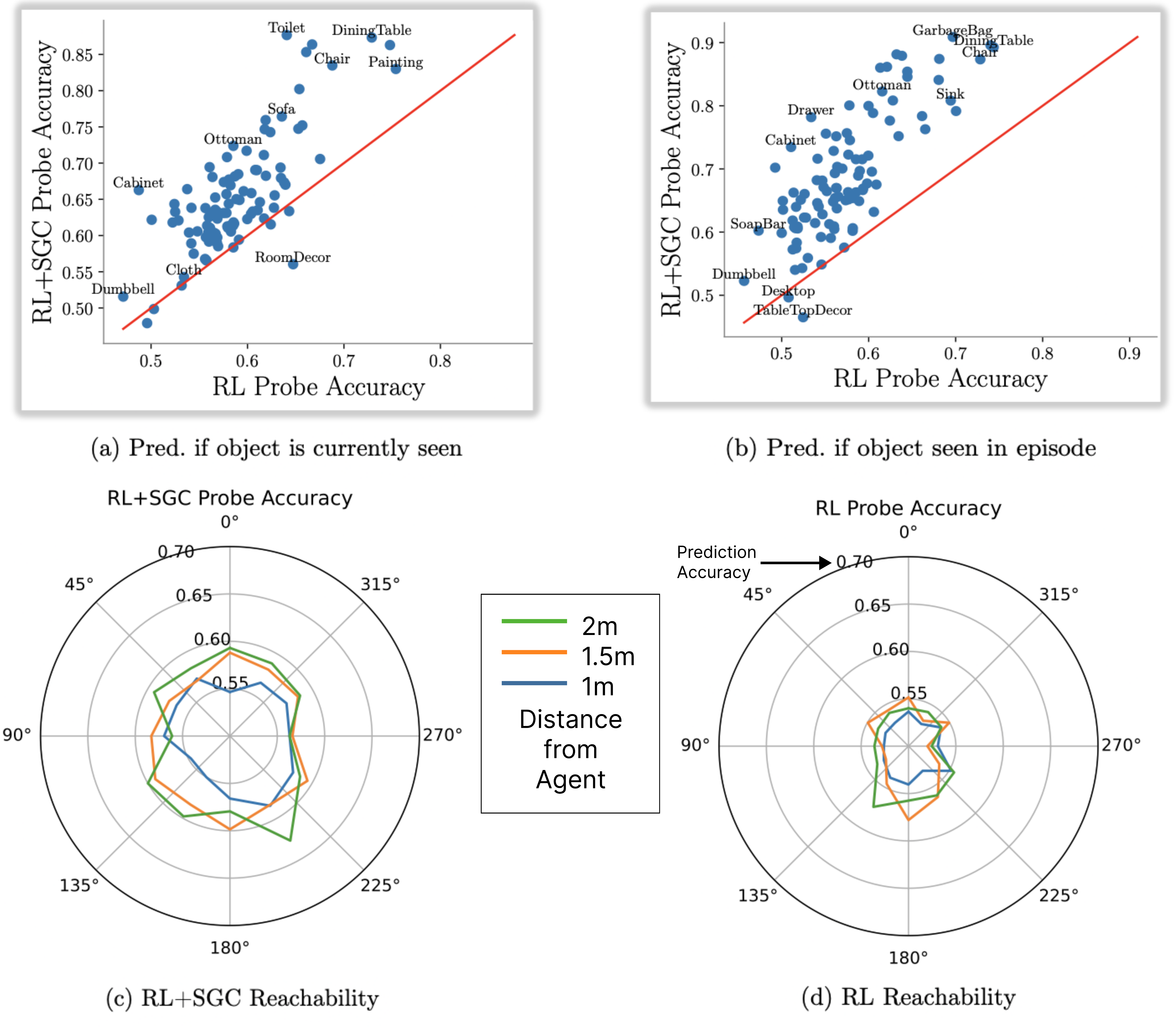}
\caption{\textbf{Linear probing}. (a) Balanced accuracy of predicting if a given object is currently visible to the agent. (b) As for (a) but predicting if an object was ever seen till that point. (c) and (d) denote the accuracy for predicting if a location is reachable by the agent for RL{$+$}SGC and RL methods respectively. In \{\textcolor{theblue}{blue}, \textcolor{theorange}{orange}, \textcolor{thegreen}{green}\}, we denote the accuracy along different directions for radii of \{1m, 1.5m, 2m\} around the agent oriented towards $0^{\circ}$.}
\vspace{-1.5em}
\label{fig:linear-probes}
\end{figure}
\subsubsection{Probing Learned Representations}
To understand how SGC impacts the representations learned by our agent, we perform two linear probing experiments. Specifically, we evaluate two ObjectNav agents, our trained RL${+}$SGC and RL agents, along fixed trajectories set in the ProcTHOR training scenes. At each step, we save both the agent's current belief states and additional metadata regarding what areas around the agent are free-space and what objects are visible to the agent. We then partition this data into training, validation, and testing splits, and train linear probes upon the frozen agent beliefs to predict the saved metadata. In particular, we fit binary logistic regression models to predict, for each object category: whether or not that object is currently visible and, to test agent memory, whether or not the agent has seen the object previously during the episode. We also fit such models to predict, at every step, whether or not various locations around the agent are ``free-space'' (\ie can be occupied by the agent without collision). We summarize our test-set results in Fig.~\ref{fig:linear-probes}. We find that the RL${+}$SGC agent is, almost uniformly across object categories, better able to both predict which categories are visible and if they were previously seen. Similarly, the RL${+}$SGC agent's beliefs are uniformly better at predicting free-space about the agent, especially when predicting free-space \emph{behind} the agent. Together this suggests that the SGC loss encourages developing both a semantic and geometric understanding of the environment. For further details, see our supplementary materials.

\subsubsection{Importance of positional information}
As discussed in Section~\ref{sec:scene_graph_build}, we encode the 3-D spatial positions of objects relative to the agent at each node in the graph. A knowledge of spatial positions can allow the agent to disambiguate between object instances of the same type, and enable the agent to spatially locate objects it had seen at previous time steps.
To investigate this the importance of this positional information, we remove it from our scene graph and train an ObjectNav model which we refer to as RL${+}$SGC \emph{without position}. 
We observe that masking out position features achieves 48.4\% success rate and 28.4\% SPL on RoboTHOR object navigation. This 4\% drop in performance suggests that encoding the positional information within the graph likely enables the agent to have better spatial awareness about its own state and the objects that it has seen, and thus leads to better performance at ObjectNav.

%% file: 05_conclusion.tex
\section{Conclusion.}
We propose Scene Graph Contrastive~(SGC) learning as a \emph{general-purpose}, \emph{supervisory} signal for training embodied agents. SGC employs non-parametric scene graphs as a \emph{training-only} signal in a contrastive learning framework. It effectively asks the agent to pick-out the graph corresponding to its present and past observations, thereby encouraging the agent to develop a graph-aware belief state. We evaluate SGC, by training agents on three embodied tasks, Object Navigation, Multi-Object Navigation and Arm Point Navigation and show performance improvements across all of them. Additionally, we evaluate the quality of our belief representations by showing adaptation to novel object categories and via a linear probing analysis.

\noindent\textbf{Limitations and future work.}
One key limitation of our approach is that building an iterative scene graph comes with a computational overhead which increases training time for embodied agents. This stems from metadata generation required for graph building, and calculating the relationships in the scene graph. Engineering solutions to speed these up can allow denser sampling of graphs and lead to a potentially stronger training signal.  
We use vanilla Graph Attention Networks~\cite{Velickovic2018GraphAN} to encode scene graphs. Stronger graph encoders models have been proposed and these may provide an even richer graph representation and lead to better embodied agents. 

%% file: 06_appendix.tex
\section{Scene Graph Details.}

\subsection{Details about relationships in scene graphs.}
As mentioned in Section 3.1, we use the edges of the scene graph to encode various kinds of relationships. Note that all these relationships are directed, therefore are not commutative in nature. We define three \textbf{Agent-Object} relationships, namely \emph{Sees}, \emph{Holds} and \emph{Touches}. \emph{Sees} encodes whether an object is currently within 2.5 meters and in the field of view of the egocentric camera on the agent. \emph{Holds} indicates whether an object is picked up by the agent. We only use it with the arm based agent~\cite{manipulathor} for ArmPointNav. Similarly, \emph{Touches} is also applicable when using an arm-based agents. It includes objects that the agent has picked up, and also the ones that its arm might have collided with. 

We have three \textbf{Object-Object} relationships, namely \emph{On}, \emph{Near} and \emph{Adjacent}. An object is considered \emph{Near} another object, if the distance between them is below a certain threshold. However, for an object to be \emph{Adjacent} to another, they need to be \emph{Near} and also have an unobstructed path between their centers.
We also have \textbf{Agent-Conditioned Object-Object} spatial relationships that include \emph{Right}, \emph{Left}, \emph{Above}, \emph{Below}, \emph{Front}, and \emph{Behind}. These relationships are determined based on the each object's position relative to the agent's coordinate frame.
Lastly we define a relationship \emph{Contains} which can be an Object-Object, Room-Object or a Room-Agent relationship. For instance, if the agent is present in the kitchen, the relationship ``Kitchen \emph{Contains} Agent" would be true. 

To summarise, we define the following relationships between various nodes of the graph:\\

\newcommand{\tabitem}{~~\llap{\textbullet}~~}
\begin{tabular}{llll}
\tabitem Sees & \tabitem Touches & \tabitem Holds & \tabitem On \\ \tabitem Near & \tabitem Adjacent & \tabitem Right & \tabitem Left \\ \tabitem Above & \tabitem Below & \tabitem Front & \tabitem Behind \\ \tabitem Contains \\
\end{tabular}

\subsection{Importance of retaining history.}
Section 3.1 discusses how we build an iterative scene graph from the agent's exploration of the environment. Once an object is added as a node to graph, it continues to exist in the graph, even if it goes out of view. We update the relationships and node position features between each pair of nodes at every step. One disadvantage of retaining the history of nodes is increasingly larger scene graphs as the episode progresses. This leads to some computational and memory overhead which lead us to investigate the importance of preserving the history in the scene graph. We train a model with a variant of the SGC loss that removes the history of nodes, thereby constructing a graph with only objects that are visible at the current time step. We refer to this model as RL ${+}$ SGC-\emph{no hist}.  

We summarize the results in Table~\ref{tab:no_history}. We find that when SGC is trained without retaining the history of nodes, it ends up performing worse than just RL. We believe that SGC-\emph{no hist.}~would encourage the agent's belief representation to only remember information about its observations at a given time step. This form of short term memory harms its exploration abilities and leads it to traverse much shorter trajectories, as reflected by an episode length of 126, which is significantly lower than RL and RL ${+}$ SGC variants. Additionally, RL ${+}$ SGC-\emph{no hist.}~also takes significantly higher \emph{False End} actions, which essentially indicates it terminates the episode without finding the target object much more often than the other models.    

\begin{table}[t]
\centering
\begin{tabular}{lcccc}
\toprule
Model                    & SR   & SPL  & EpLen & False End \\ \midrule
RL (baseline)                      & 41.0   & 28.0   & 193   & 45.0\%      \\
RL + SGC (ours)                & 53.4 & 32.8 & 245   & 14.9\%    \\
RL + SGC-\emph{no hist.} & 34.8 & 26.1 & 126   & 60.6\%  \\ \bottomrule
\end{tabular}
\caption{\textbf{Metrics on RoboTHOR ObjectNav benchmark.} EpLen denotes the average episode length traversed by the agent. False End denotes the amount of \textsc{End} actions that resulted in failure.}
\label{tab:no_history}
\end{table}

\section{Experiment Details.}
\begin{table*}[t!]
\centering
\begin{tabular}{lccc}
\toprule
Object Type  & RoboTHOR & AI2-iTHOR & \textsc{ArchitecTHOR} \\ \midrule
Alarm Clock  &  \greencheck        &  \greencheck         &   \greencheck           \\
Apple        &  \greencheck        & \greencheck           &    \greencheck          \\
Baseball Bat &  \greencheck        & \greencheck           &    \greencheck          \\
Basketball   &  \greencheck        & \greencheck           &    \greencheck          \\
Bed          &  \redcross      &  \greencheck         & \greencheck              \\
Bowl         &  \greencheck        &    \greencheck       & \greencheck             \\
Chair        &   \redcross       &   \greencheck        & \greencheck              \\
Garbage Can  &  \greencheck        &   \greencheck        & \greencheck             \\
House Plant  &  \greencheck        &    \greencheck       & \greencheck             \\
Laptop       &  \greencheck        &    \greencheck       & \greencheck             \\
Mug          &  \greencheck        &    \greencheck       & \greencheck             \\
Sofa         &  \redcross        &   \greencheck        & \greencheck              \\
Spray Bottle &  \greencheck        & \greencheck          & \greencheck             \\
Television   &  \greencheck        & \greencheck          & \greencheck             \\
Toilet       &   \redcross       & \greencheck          & \greencheck              \\
Vase         &  \greencheck        &  \greencheck         & \greencheck             \\ \bottomrule
\end{tabular}
\caption{\textbf{Target object types used for each ObjectNav benchmark.}}
\label{tab:object_types}
\end{table*}

\noindent In this section, we provide the training and hyperparameter settings for all our models on Object Navigation~(ObjNav), Multi-Object Navigation~(MultiON) and ArmPointNav. 

\subsection{ObjectNav.}
For ObjectNav, an agent is tasked to find a target object type (e.g.~bed) in an environment. We provide the target object type in the form of an embedding, and use forward-facing egocentric RGB images at each time step. All our ObjectNav models are trained with a simulated LoCoBot~\cite{locobot} agent. We describe the task and training details below.

\noindent\textbf{Evaluation}. Following~\cite{Anderson2018OnEO}, an ObjectNav task is considered successful if:

\noindent $\bullet$ The agent terminates the episode by calling the \textsc{End} action.

\noindent $\bullet$ An instance of the target object type is within a distance of 1 meter from the agent.

\noindent $\bullet$ The object is visible from the agent's camera. If the object is occluded behind an obstacle or out of the agent's view, the episode is considered unsuccessful.

We also use SPL~\cite{Anderson2018OnEO,Batra2020ObjectNavRO} to evaluate the efficiency of our agents. If an environment has multiple instances of the target object type, if the agent navigates to any of those instances, it is considered successful. For calculating the SPL in such scenarios, the shortest path is defined as the minimum shortest path length from the starting position to any of the reachable instance of the given type, regardless of which instance the agent navigates towards.

We evaluate \emph{zero-shot} on RoboTHOR~\cite{robothor}, AI2-iTHOR~\cite{ai2thor} and \textsc{ArchitecTHOR}~\cite{procthor} benchmarks. To reiterate, \emph{zero-shot} here implies that no training was performed on any of these benchmarks. Table~\ref{tab:object_types} shows the object types used for respective benchmarks. We use the PRIOR package~\cite{prior} for loading datasets for all benchmarks.

\noindent\textbf{Training.}
Each agent is trained using DD-PPO~\cite{Wijmans2020DDPPO}, using a clip parameter $\epsilon = 0.1$, an entropy loss coefficient of 0.01, and a value loss coefficient of 0.5. We use the reward structure and model architecture from~\cite{Khandelwal2022SimpleBE,procthor}. We summarize the training hyperparameters in Table~\ref{tab:hyperparam_objectnav}.

\begin{table}[t!]
\centering
\begin{tabular}{ll}
\toprule
Hyperparameter           & Value \\ \midrule
Discount factor ($\gamma$)  & 0.99  \\
GAE~\cite{Schulman2016HighDimensionalCC} parameter ($\lambda$)   & 0.95  \\
Value loss coefficient   & 0.5   \\
Entropy loss coefficient & 0.01  \\
Clip parameter ($\epsilon$) & 0.1   \\
Rollout timesteps        & 20    \\
Rollouts per minibatch   & 1     \\
Learning rate            & 3e-4  \\
Optimizer                & Adam~\cite{Kingma2015AdamAM}  \\
Gradient clip norm       & 0.5   \\ \bottomrule
\end{tabular}
\caption{\textbf{Training hyperparameters for ObjectNav.}}
\label{tab:hyperparam_objectnav}
\end{table}

\begin{table}[t!]
\centering
\begin{tabular}{ll}
\toprule
Hyperparameter           & Value \\ \midrule
Discount factor ($\gamma$)  & 0.99  \\
GAE~\cite{Schulman2016HighDimensionalCC} parameter ($\lambda$)   & 0.95  \\
Number of RNN Layers   & 1   \\
Step penalty  & -0.01  \\
Gradient Steps & 128   \\
Rollouts per minibatch   & 1     \\
Learning rate            & 3e-4  \\
Optimizer                & Adam~\cite{Kingma2015AdamAM}  \\
Gradient clip norm       & 0.5   \\ \bottomrule
\end{tabular}
\caption{\textbf{Training hyperparameters for ArmPointNav.}}
\vspace{-1em}
\label{tab:hyperparam_armpointnav}
\end{table}

As mentioned in the main paper, we train our ObjectNav agents in \emph{ProcTHOR-train} environments from the ProcTHOR-10k dataset~\cite{procthor}. We train with 16 target object types shown in Table~\ref{tab:object_types}. We train with 48 parallel processes on 8 NVIDIA RTX A6000 GPUs for 350 million steps. All the ObjectNav models we present took 6-8 days to train depending on whether they had an auxiliary objective or not. We follow the strategy proposed in~\cite{procthor} to sample target object types during training. 

\subsection{Multi-ObjectNav.}

In Multi-ObjectNav~(MultiON), an agent needs to find multiple target objects in a particular order. We perform experiments with two variants of this task, MultiON-2 and MultiON-3 that requires an agent to find 2 and 3 target objects in an episode, respectively.

\noindent\textbf{Evaluation.}
We collect a validation set in 200 \emph{ProcTHOR-Val} houses using the same set of target objects presented in Table~\ref{tab:object_types}. The agent needs to issue a \textsc{Found} action to indicate if it found the requested object. We build upon the success criteria for ObjectNav to define a successful \textsc{Found} action as:

\noindent $\bullet$ An instance of the current target object type is within a distance of 1 meter from the agent.

\noindent $\bullet$ The current target object is visible from the agent's camera. If the target object is occluded or out of the agent's view, the \textsc{Found} action is considered unsuccessful.

For a MultiON-2 episode to be successful, the agent requires to execute 2 successful \textsc{Found} actions. A failed \textsc{Found} action at any point in the episode renders it as a failure. We extend the same criteria to 3 objects for MultiON-3.

\noindent\textbf{Training.}
Similar to ObjectNav, we train in \emph{ProcTHOR-train} environments from the ProcTHOR-10k dataset~\cite{procthor}. We use the 16 target object types presented in Table~\ref{tab:object_types}, and sample 2 or 3 object types for each episode based on the Multi-ON variant. We extend the object sampling strategy from~\cite{procthor} to multiple objects to ensure roughly uniform sampling of target object types. At the beginning of each episode, we provide the first target object, and provide the next target object only after a successful \textsc{Found} action. This allows us to use the EmbodiedCLIP~\cite{Khandelwal2022SimpleBE} Object Navigation architecture. We use the same distance based reward shaping from~\cite{procthor,Khandelwal2022SimpleBE}, and provide a reward of 10.0 for each successful \textsc{Found} action. 

Each MultiON agent is trained with DD-PPO~\cite{Wijmans2020DDPPO} using the same set of hyperparameters as ObjectNav presented in Table~\ref{tab:hyperparam_objectnav}. We train with 48 parallel processes on 8 NVIDIA RTX A6000 GPUs for 180 million steps. Training takes 2 days with just RL and 5 days with the RL ${+}$ SGC baseline.

\subsection{ArmPointNav.}
For ArmPointNav, we follow the same architecture as~\cite{manipulathor} with the exception of using a CLIP-pretrained frozen ResNet-50 encoder as our visual backbone instead of a 3-layer CNN. Our motivation for this design choice was improved results for the RL model with the CLIP backbone. Therefore, we use CLIP-pretrained ResNet-50 for both RL and RL ${+}$ SGC models. 

\noindent\textbf{Evaluation.} 
We evaluate our model on the validation set from~\cite{manipulathor}. The task requires the agent to move a target object from a starting location to a goal location using the relative location of the target in the agent's coordinate. It only uses egocentric RGB observation as its visual input. 

\noindent\textbf{Training.}
Following~\cite{procthor}, we train our ArmPointNav models on a subset of 7,000 houses with 58 object categories. For each training episode, we teleport the agent to a random location, randomly sample a pickupable object, and randomly sample a target location. We train our models for 90 million steps on 8 NVIDIA RTX A6000s. We list the training hyperparameters in Table~\ref{tab:hyperparam_armpointnav}. Training takes 80 hours with just RL and 100 hours with RL ${+}$ SGC. We note that SGC is applied to 20\% of the steps in a batch.

\subsection{Adapting to Novel Object Categories.}
As presented in Section~4.5.2 in the main draft, we show that models trained with SGC are able to adapt to novel object categories much faster. We provide the 10 sets of 5 Object Categories sampled for this experiment in Table~\ref{tab:novel_object_sets}. We also provide the larger pool of 80+ categories that they were sampled from in Table~\ref{tab:all_obj_types}.

\begin{table}[t]
\centering
\resizebox{\columnwidth}{!}{%
\begin{tabular}{ll}
\toprule
Set1  & Pot, Egg, CellPhone, CD, ToiletPaper                   \\
Set2  & CellPhone, Faucet, Fork, Desktop, Box                  \\
Set3  & Newspaper, Lettuce, ButterKnife, Spatula, CellPhone    \\
Set4  & Lettuce, Faucet, CoffeeMachine, Pot, Stool             \\
Set5  & TeddyBear, SideTable, Lettuce, DeskLamp, CoffeeMachine \\
Set6  & TVStand, Safe, Desktop, Ottoman, Tomato                \\
Set7  & WashingMachine, Toaster, SideTable, Plunger, Desk      \\
Set8  & Drawer, Statue, Toaster, Fridge, Newspaper             \\
Set9  & Microwave, FloorLamp, Sink, DiningTable, Stool         \\
Set10 & Drawer, SinkBasin, LaundryHamper, DeskLamp, Sink       \\ \bottomrule
\end{tabular}%
}
\caption{\textbf{Sets of randomly sampled object categories.}}
\label{tab:novel_object_sets}
\end{table}

\begin{table*}[t!]
\centering
\begin{tabular}{l}
\toprule
ArmChair, Book, Boots, Bottle, Box, Bread, ButterKnife, CD, Cabinet, Candle, Cart, Cellphone, Cloth,\\ClothesDryer, CoffeeMachine, CoffeeTable, CounterTop, CreditCard, Cup, Desk, DeskLamp,
Desktop,\\DiningTable, DishSponge, DogBed, Doorframe, Doorway, Drawer, Dresser, Dumbbell, Egg, Faucet, Floor,\\FloorLamp, Fork, Fridge, GarbageBag, Kettle, KeyChain, Knife, Ladle, LaundryHamper, Lettuce, Microwave,\\Newspaper, Ottoman, Painting, Pan, PaperTowelRoll, Pen, Pencil, PepperShaker, Pillow, Plate, Plunger, Pot,\\Potato, RemoteControl, RoomDecor, Safe, SaltShaker, Shelf, ShelvingUnit, SideTable, Sink, SinkBasin, SoapBar,\\SoapBottle, Spatula, Spoon, Statue, Stool, TVStand, TableTopDecor, TeddyBear, TennisRacket, TissueBox,\\Toaster, ToiletPaper, Tomato, VacuumCleaner, Wall, WashingMachine, Watch, Window, WineBottle\\
\bottomrule
\end{tabular}%
\caption{\textbf{List of All Object Types.}}
\label{tab:all_obj_types}
\end{table*}

\section{Linear Probing Details.}

\noindent\textbf{Data capture.} We collect data from about 5 episodes in each of approx.~2,500 ProcTHOR-10K~\cite{procthor} houses, using an exploration policy that iteratively visits the nearest unvisited reachable object in the house until no more objects are left to visit. During the trajectory, we store the current scene graph, both RL and RL ${+}$ SGC agents' beliefs, and environment metadata like agent pose and scene name.

\noindent\textbf{Processing pipeline.} For each probing experiment, given the collected beliefs and task-dependent binary labels, we subsample the collected data to produce a compact set of training samples. In order to improve numerical stability, we compress the beliefs via PCA, ensuring 99\% of the data variance is explained. We fit a logistic regression model on the compressed training dataset, and evaluate on an (also compressed) held-out test set. The used implementations for PCA and logistic regression are from \cite{scikit-learn}.

\noindent\textbf{Reachability.} For these experiments, we define reachability as whether a location at an Euclidean distance $R$ from the agent and with relative orientation $\theta$ is ``free-space'' (\ie, can be occupied by the agent without collision).
The free space of a scene is estimated by applying marching squares \cite{lorensen1987,scikit-image} with a threshold level of 0.9999 to its binary grid of reachability (with \{0, 1\} indicating \{occupied, free\} space).
We consider three concentric circumferences around the agent's current location and locate points on these circumferences that are at $30^{\circ}$ spaced angles with respect to the agent's current orientation, resulting in 36 unique relative locations. Each of the 36 relative locations is used as a target variable.
In these experiments we use 20,000 data points for each of the 36 target variables, of which 18,000 are for training and 2,000 for testing. We sample these data points so that negative and positive samples are balanced for each target variable, both for training and for testing. All accuracy metrics provided for this task are thus ``balanced accuracy" rather than standard accuracy.
We fit the 36 models using the processing pipeline described above.

\noindent\textbf{Visibility.} In these experiments we subsample 40,000 datapoints of which 33,000 are used for training, 2,000 for validation, and 5,000 for testing. In order to estimate an agent's understanding of visibility, we record, at every sampled location, which objects where visible to the agent within a 2.5 meters range. As most objects will not be visible at any given agent position, na\"{i}vely sampling the 40,000 datapoints above would result in many objects having almost no positive (\ie visible) examples. Because of this, we use an iterative sampling procedure that prioritizes selecting samples which contain instances of visible objects that are otherwise underrepresented in the dataset. Even using this iterative approach there is significant class imbalance with objects of type \textsc{RoomDecor} being visible in only 2\% of datapoints (the minimum across all object types) while the \textsc{Painting} objects are visible in 20\% (the maximum across all object types). For this reason, when we fit separate logistic regression models to predict visibility for each of these object types we reweigh samples so as to ensure that the negative and positive examples have, in total, equal importance. When computing accuracy metrics we thus also use the ``balanced accuracy'' rather than the standard accuracy which would overweigh negative examples in this setting. In the main paper we also report results when fitting models exactly as above but using, as target, whether or not an object was visible at the current timestep or any previous time step in the episode so far. These additional results give insight into the agents' ability to remember having seen objects in the past.

\noindent\textbf{Revisited state detection.} In addition to the two probing experiments described in the main paper, we also probe for the detection of revisited states. We quantize the current agent pose relative to the one at the beginning of the episode using square cells of $0.25\times 0.25~{m^2}$ for location and arcs of 30 degrees for rotation. We treat each pair of relative grid location and rotation as a separate state. In these experiments we use 20,000 data points, of which 18,000 are for training and 2,000 for testing. The RL ${+}$ SGC agent achieves 56.20\% balanced accuracy, whereas the RL-only agent reaches 52.10\% on this task.

\begin{figure}[t!]
\includegraphics[width=0.5\textwidth]{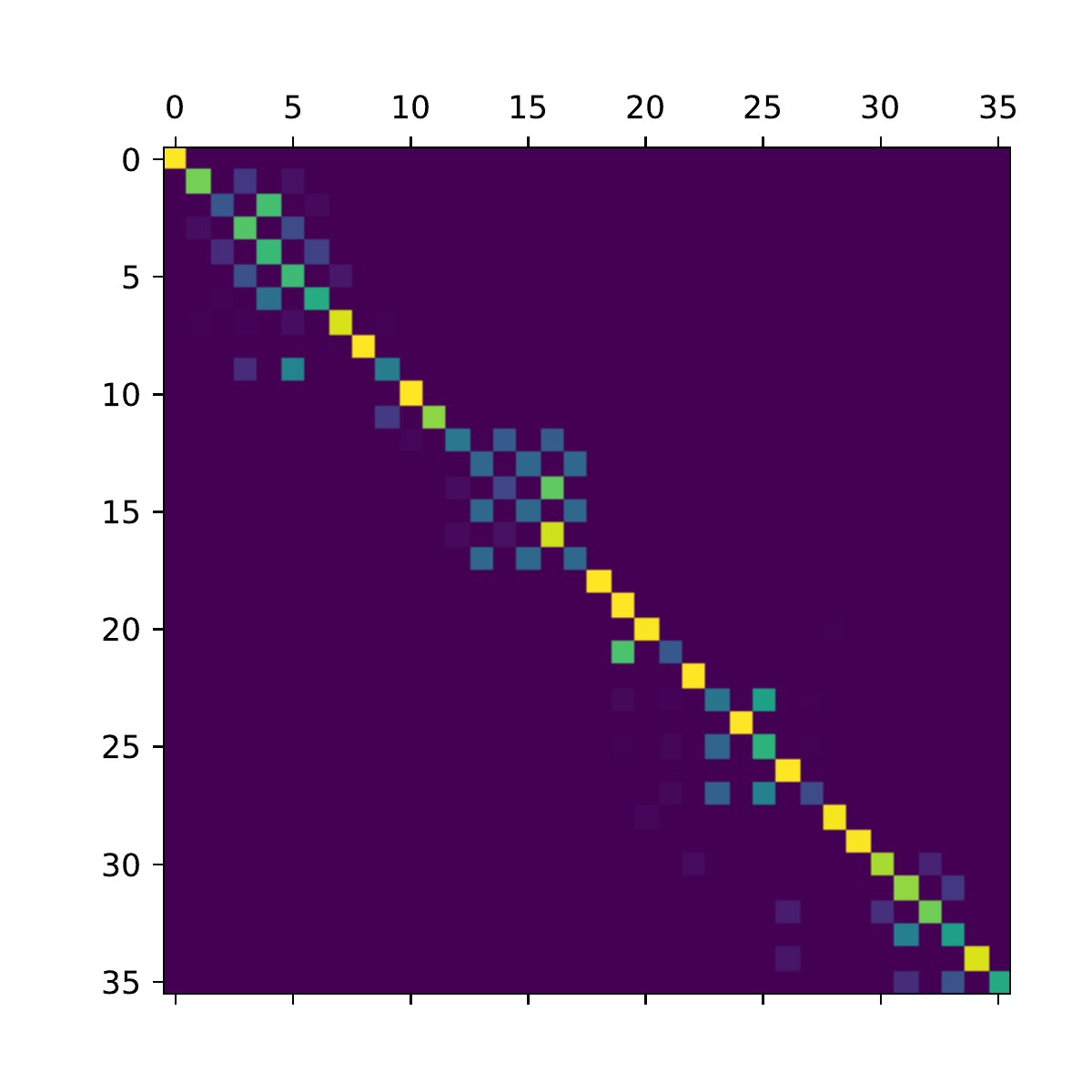}
\caption{\textbf{Graph Loss Plot.}}
\label{fig:loss_plot}
\end{figure}

\section{ObjectNav Qualitative Analysis.}
We visualize some trajectories and provide qualitative analysis on how SGC is affecting the ObjectNav behavior. We attach the videos and trajectory maps with the supplementary zip file. Each trajectory map shows the top down view of the path traversed by the agent. It also indicates the target object location by a purple box. We indicate the target object type and the number of steps on the bottom left corner of the frame in the trajectory videos.  

\paragraph{\textsc{ArchitecTHOR.}}
Firstly we show some top-down trajectories in \textsc{ArchitecTHOR} environments (Fig.~\ref{fig:architecthor-qualitative}), noting the floor plans' large scale and photorealism.
Example 1 shows the ObjectNav agents trying to find a spray bottle. The RL-only model looks into one part of the house, then terminates the trajectory unsuccessfully, whereas RL ${+}$ SGC is able to navigate to the bathroom and successfully finds the target. 

Example 2 shows an example where the RL-only agent falsely recognizes the microwave as a television. The RL ${+}$ SGC agent also starts by checking the kitchen and the microwave, then turns around and explores another region and is able to find the television successfully. 

Example 3 shows an interesting failure case for both RL and RL ${+}$ SGC models.
The target object is a vase. The RL-only agent looks for the vase in the area around the dining table and then terminates the trajectory failing to find the target. The trajectory map shows the limited exploration of this model.

On the other hand, even though the RL ${+}$ SGC agent fails to find the target, it explores the room in a much more exhaustive fashion as can be seen in the respective trajectory map. It exhausts the maximum number of steps allowed and times out instead of falsely calling an \textsc{End} action.

\newcommand{\vertscale}{0.34\linewidth}

\begin{figure*}
    \centering
    \begin{tabular}{cc}
        RL-only & RL ${+}$ SGC \\
        \\
        \frame{\includegraphics[height=\vertscale]{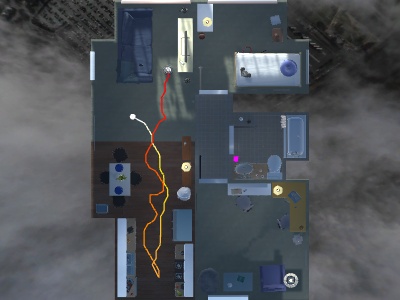}}&
        \frame{\includegraphics[height=\vertscale]{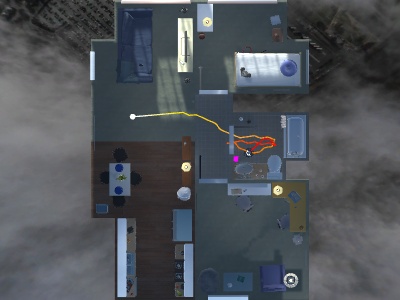}}\\
        \multicolumn{2}{c}{Example 1. Target object type: Spray bottle.}\\
        \\
        \frame{\includegraphics[height=\vertscale]{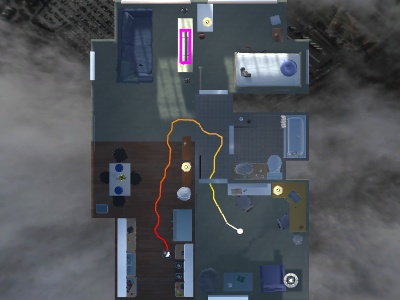}}&
        \frame{\includegraphics[height=\vertscale]{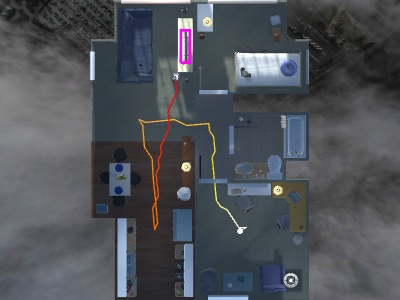}}\\
        \multicolumn{2}{c}{Example 2. Target object type: Television.}\\
        \\
        \frame{\includegraphics[height=\vertscale]{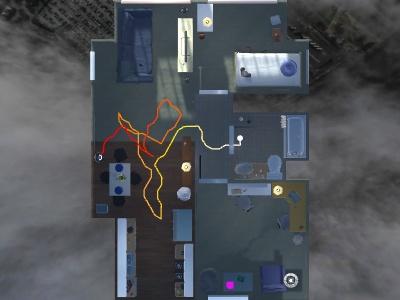}}&
        \frame{\includegraphics[height=\vertscale]{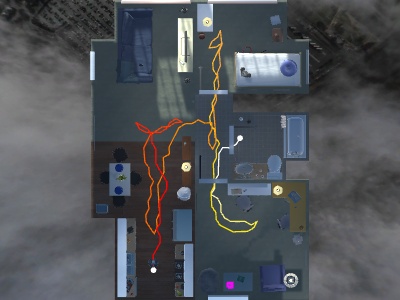}}\\
        \multicolumn{2}{c}{Example 3. Target object type: Vase.}\\
    \end{tabular}
    \caption{\label{fig:architecthor-qualitative}\textbf{Top-down trajectories for RL-only and RL ${+}$ SGC agents in \textsc{ArchitecTHOR}.} Examples 1 and 2 correspond to successful trajectories for the RL ${+}$ SGC agent, whereas Example 3 shows different behaviors for failed episodes. The target object type in each episode is highlighted in magenta. For the trajectory lines, \{white, red\} correspond to the \{beginning, end\} of each episode.}
\end{figure*}

\begin{figure*}
    \centering
    \begin{tabular}{cc}
        RL-only & RL ${+}$ SGC \\
        \\
        \frame{\includegraphics[height=\vertscale]{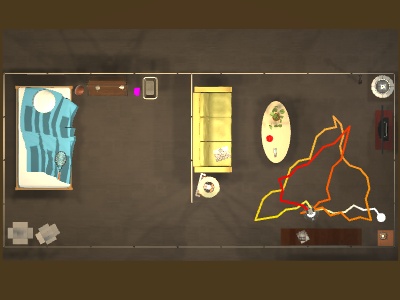}}&
        \frame{\includegraphics[height=\vertscale]{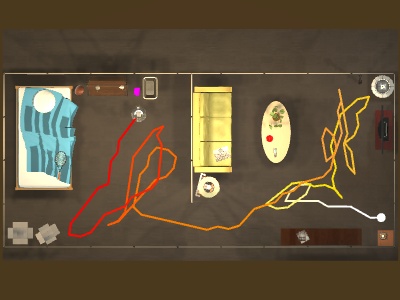}}\\
        \multicolumn{2}{c}{Example 1. Target object type: Spray bottle.}\\
        \\
        \frame{\includegraphics[height=\vertscale]{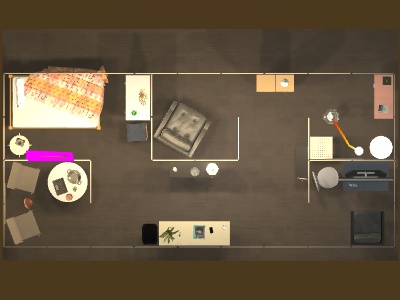}}&
        \frame{\includegraphics[height=\vertscale]{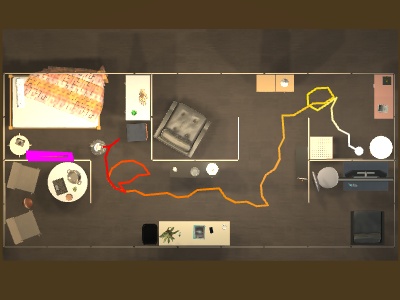}}\\
        \multicolumn{2}{c}{Example 2. Target object type: Baseball bat.}\\
        \\
        \frame{\includegraphics[height=\vertscale]{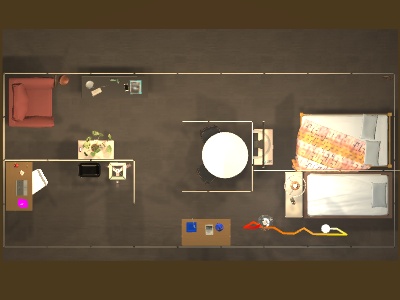}}&
        \frame{\includegraphics[height=\vertscale]{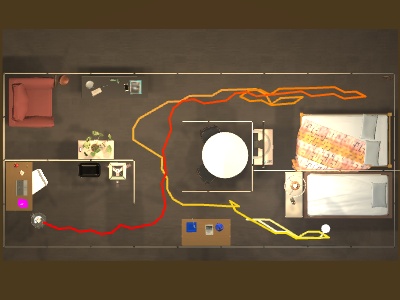}}\\
        \multicolumn{2}{c}{Example 3. Target object type: Mug.}\\
    \end{tabular}
    \caption{\label{fig:robothor-qualitative}\textbf{Top-down trajectories for RL-only and RL ${+}$ SGC agents in RoboTHOR.} The target object type in each episode is highlighted in magenta. For the trajectory lines, \{white, red\} correspond to the \{beginning, end\} of each episode.}
\end{figure*}

\paragraph{RoboTHOR.}
We show three qualitative examples from the RoboTHOR benchmark as well (Fig.~\ref{fig:robothor-qualitative}). They all seem to highlight the inability of the RL-only agent to explore the scene, which leads to unsuccessful termination of the episode. On the other hand, the RL ${+}$ SGC agent explores the region it starts in and then goes to different parts of the scene, which eventually allows it to succeed. %

\section{Loss Behavior.}
To show the SGC loss predictions, we present a plot of prediction probabilities for our ObjectNav RL + SGC model in Figure~\ref{fig:loss_plot}. We show the plot with 18 rollout steps across 2 parallel processes \ie with a total batch size of 36. Note how the predictions are concentrated around the diagonal which is akin to the ground truth present in Figure 3 of the main paper. It is also interesting to see some deviations from the diagonal, which shows that there is some scope to optimize the loss even better and potentially lead to more performance benefits. These deviations arise from scene graphs from nearby time steps in the same episode being very slightly different, making it harder to classify them.

